\documentclass{IOS-Book-Article}

\usepackage{cite}
\usepackage{amsmath,amssymb,amsfonts}
\usepackage{algorithmic}
\usepackage{graphicx}
\usepackage{textcomp}
\usepackage{hyperref}
\usepackage{mathptmx}
\usepackage{soul}\setuldepth{article}

\def\hb{\hbox to 11.5 cm{}}

\begin{document}
\pagestyle{headings}
\def\thepage{}
\begin{frontmatter}              

\title{Using Deep Reinforcement Learning for Zero Defect Smart Forging}
\markboth{}{January 2022\hb}

\author[A]{Yunpeng Ma%
\thanks{Corresponding Author: Yunpeng Ma, Karlstad Universitet, Sweden; E-mail:yunpeng.ma@kau.se, this paper is accepted in the 10th Swedish Production Symposium 2022 (SPS2022)
}
},
\author[A]{Andreas Kassler},
\author[A]{Bestoun S. Ahmed},
\author[B]{Pavel Krakhmalev},
\author[C]{Andreas Thore},
\author[D]{Arash Toyser}, and 
\author[E]{Hans Lindbäck}

\runningauthor{Yunpeng Ma et al.}
\address[A]{Computer Science Department, Karlstads Universitet, Karlstad, Sweden}
\address[B]{Materials Science Department, Karlstads Universitet, Karlstad, Sweden}
\address[C]{Smart Industrial Automation, RISE Research Institutes of Sweden, Västerås, Sweden}
\address[D]{Viking Analytics AB, Göteborg, Sweden}
\address[E]{Bharat Forge Kilsta AB, Karlskoga, Sweden}

\begin{abstract}
Defects during production may lead to material waste, which is a significant challenge for many companies as it reduces revenue and negatively impacts sustainability and the environment. An essential reason for material waste is a low degree of automation, especially in industries that currently have a low degree of digitalization, such as steel forging. Those industries typically rely on heavy and old machinery such as large induction ovens that are mostly controlled manually or using well-known recipes created by experts. However, standard recipes may fail when unforeseen events happen, such as an unplanned stop in production, which may lead to overheating and thus material degradation during the forging process. In this paper, we develop a digital twin-based optimization strategy for the heating process for a forging line to automate the development of an optimal control policy that adjusts the power for the heating coils in an induction oven based on temperature data observed from pyrometers. We design a digital twin-based deep reinforcement learning (DTRL) framework and train two different deep reinforcement learning (DRL) models for the heating phase using a digital twin of the forging line. The twin is based on a simulator that contains a heating transfer and movement model, which is used as an environment for the DRL training. Our evaluation shows that both models significantly reduce the temperature unevenness and can help to automate the traditional heating process.
\end{abstract}

\end{frontmatter}
\markboth{January 2022\hb}{January 2022\hb}
\section{Introduction }
For many companies, material waste during production is a significant problem. It reduces revenue and negatively impacts sustainability and the environment. An essential reason for defects during production that causes material waste is a low degree of automation. Especially, many heavy and more traditional industries, including steel and metal processing, still rely on existing equipment that is not fully automated. Instead, the production relies on the manual operation or semi-automated production solutions using existing recipes that have been created by experts in the field long ago. However, such recipes typically do not cover all events during production and are relatively simple. In addition, they frequently may need to be overridden manually by the interaction of a skilled operator to minimize material wastage caused by non-optimal production processes. A prominent example is the scrap rate caused by the steel quality degradation in the induction heating process for a forging line. During production, root causes for defects are rapid and imprecise manual adjustments of the suggested power to the induction coils that heat the steel beams and the roller conveyors that transport the beams to the forging press. 

During recent years, the manufacturing industry has seen a constant need for improving production processes, reducing process failures, and enhancing the quality of products aiming towards zero-defect manufacturing (ZDM)\cite{Psarommatis2021ZDM}. The goal of ZDM is to minimize and eliminate defects by improving control systems and using nondestructive inspection methods. In turn, it will significantly benefit from digitalization and process automation \cite{Mourtzis2016bigdata, Serrano-Ruiz2021smartmanufacturing, Psarommatis2021ZDM}. The recent push towards the fourth industrial revolution (Industry 4.0)\cite{Qin2016industry4.0} aims to increase digitalization and automation for the traditional industry by pushing key enabling technologies, such as machine learning and artificial intelligence (AI), data analytics, digital twins, cloud and edge computing, and industrial internet-of-things (I-IoT) to the shop floor. However, adopting those technologies to achieve ZDM in heavy industries is challenging as traditional processes are not fully digitized. In addition, the potential of machine learning and AI for optimizing production processes relies on digital twins, which aim to build a realistic model of the process and the product in a simulator that can be used to simulate different production strategies.

In this paper, we aim to achieve ZDM by optimizing the heating process in the forging line. This is challenging as the process has several hard requirements that must be maintained during the production process so that the steel has the desired quality. Our approach builds a digital twin-based deep reinforcement learning (DTRL) framework to find the optimal control policy of the heating process for the forging line to adjust the power based on temperature data observed from pyrometers mounted in the induction heaters. We first build a digital twin for the heating process, which estimates the temperature distribution over the steel bar for a given power to the induction coils in the different zones of the oven and uses different speeds for the roller conveyors. We describe how the simulator can be instrumented for movement and heat transfer simulations in normal production and warm holding mode. We also describe our optimization problem to minimize the defects due to metal under heat or overheat subject to the conveyors' movement and the heating power constraints. We use the digital twin as an environment to train a reinforcement learning agent to find an optimal control policy. As the number of states is huge, we connect the agent to a deep learning framework. In our approach, we use the proximal policy optimization (PPO) algorithm as our agent, which outperforms other online policy gradient methods on the control tasks, including simulated robotic locomotion and Atari game playing\cite{schulman2017proximal}. The digital twin generates synthetic training data and acts as an environment for the reinforcement learning agent. We also propose an architecture to integrate model transfer with an edge compute platform to deploy the model in the real production environment. Our evaluation results show that the DTRL framework can automatically find an optimal control policy without requiring expert knowledge while keeping the temperature within the required range. 

The rest of the paper is organized as follows. Section \ref{Background} introduces background and related work. Section \ref{Method} presents the problem description, our digital twin-based approach, details on the simulator and the mathematical model as well as our deep reinforcement learning agent. Section \ref{Evaluation} presents evaluation results and analyses model convergence and process quality. Finally, Section \ref{Conclusion} concludes our findings and presents future work. 

\section{Background and Related Work}\label{Background}
As a specific field of machine learning, reinforcement learning (RL) has shown significant results and progress over the recent decade \cite{Nian202010reviewRL}. Significant milestones have been AlphaGo, which uses reinforcement learning to find the optimal policy for playing Go \cite{Silver2017go}. Other works used RL in the area of computer games, robotics, self-driving cars, network scheduling, among others \cite{Xu2014RLapplication, Mekrache2021RL6G}.

Recently, the potential of using RL has been explored to tackle optimal control problems through learning expressive nonlinear models in the industrial field. RL aims to find an optimal control policy using a reward function and has been used in process control \cite{Viharos2021RLprocess}. To cope with large state spaces, traditional RL has been combined with deep learning, leading towards deep reinforcement learning (DRL) \cite{Subramanian2021DRL}. Model-free DRL is a trial-and-error approach where the RL agent repeatedly interacts with the environment. The agent is trained with the training data during the interaction, including observed states, actions, and rewards. One benefit of using the DRL approach is that computational effort is shifted to the design stage by using offline training against a simulation model, where significantly higher computational power is available\cite{Nian202010reviewRL}. The optimal control policy can be evaluated efficiently and quickly after deployment on the platform.

The performance of DRL can be affected by extrinsic factors (e.g., hyperparameters or codebases) and intrinsic factors(e.g., effects of random seeds or environmental properties) \cite{Henderson2018drlmatters}. The outcome of the learning process and the interaction with the environment, model performance, as well as the required learning time, highly dependent on the choice of hyperparameters \cite{Veloso2021hyper}. Additionally, safety is a critical bottleneck when developing a DRL algorithm for real-world problems\cite{Pohland2019safety}, which in many cases may contain hard constraints that should not be violated. As DRL randomly explores the state-action space, safety control methods should be applied to avoid taking risky actions that, in practice, may lead to constraint violation.

Bøgh proposes an adaptive controller based on deep-Q networks(DQN), which copes with an HVAC system consisting of slow thermodynamics\cite{Bogh2019hvac}. DQN based algorithms are sample efficient because the transition trajectories are stored in a replay buffer used for model training. However, DQN is typically used for discrete action spaces and the optimal policy is deterministic. It cannot be used for continuous action space applications and it is not robust enough. Deep deterministic policy gradient (DDPG) is promising for continuous action space problems. For example, it has been used in the operation of gas turbines, where the control system must tackle the non-linearity and adapt to the frequent variations of operating conditions. \cite{Zhou2021gasrl} has shown that DDPG can perform better than the proportion integral differential (PID) control. PID control has a fundamental limitation imposed by the instrument, designing a robust controller to uncertain sensor signals is not an easy task and the control system may suffer from large magnitude disturbance \cite{Jorge2021PID}. On the other hand, DDPG may suffer from instability, high sensitivity to hyper-parameters, may converge to poor solutions, or not converge at all \cite{Matheron2019problem}. The proximal policy optimization (PPO) algorithm has been proposed in \cite{schulman2017proximal}, which performs comparably or better than state-of-the-art approaches while being much simpler to implement and tune\cite{Brockman2016gym}. 

In our paper, we use the PPO algorithm together with a digital twin of the heating process for the forging line to create training data so that the agent can find the optimal control policy for the industrial process. This is different from other applications such as robotic locomotion because our problem has strict hard constraints that the process needs to obey. Any violation of the constraints can cause material waste and devastate production. To the best of our knowledge, PPO has never been used for induction heating process optimization, which has different modes depending on the status of the production process. Finally, we illustrate how our architecture incorporates model transfer to deploy it on a real edge-platform connected to an industrial OPC-UA server that can interact with the real production environment.

\section{Towards Digital Twin-based Smart Forging}\label{Method}
In this section, we first describe the industrial problem setup and requirements, which resembles an induction heating process for a forging line. We introduce the main idea of using a digital twin and a deep reinforcement learning agent to find the optimal control policy. We describe the heating model used by the digital twin and its implementation in a simulator and sketch how the model can be used in a practical setup. We finish the section by describing our deep reinforcement learning model, its states, actions, and reward function. 

\subsection{Problem description and requirements} \label{3.1}

Minimizing energy and material costs as well as increasing environmental sustainability is essential for the forging process. Energy costs for heating are high due to the high amount of energy needed. The furnace is divided into five electrical induction heater zones consisting of four induction coils each. The first zone has two distinct power levels. Frequency converters with variable output voltages power the last four zones. The temperature is monitored between each coil by pyrometers. Today, the bar induction heater is manually operated. The heating result, which is essential for the process quality, is dependent on operator skill. The heating process has been tuned manually for the forging line's standard operational condition, resulting in recipes and proper settings for the induction heating coils. However, when there is a disturbance downstream in the production line, the furnace must be set in hold mode and outputting heated pieces stops. Instead, the operator starts to slowly oscillate the bar back and forward inside the heating coils. During this hold mode, the induced power is reduced not to overheat the bars. However, each heating zone has different physical characteristics, which generates an uneven temperature profile during hold mode. After a period of hold mode, the bar is more or less unevenly heated. Restarting to normal mode is challenging for the operator to adjust the converter output voltage to achieve a perfect temperature on the following billets. This task is challenging to perform manually. The result is that some outputted billets have to be scrapped because of too low temperature. Some billets are overheated and also have to be scrapped. The problem of keeping a consistent heating temperature always occurs in proximity after a hold mode period. When all materials inside the furnace are delivered to the production line, the temperature variations are back to normal and within specification without huge deviations. All scrapped material has to be replaced with new material that has to be heated to the forging temperature. A scrapped piece will result in energy and material wastage. That, in turn, has a huge impact on costs and environmental load and thus should be avoided.

We aim to eliminate the number of scrapped pieces by developing an AI-based control algorithm to find an optimal control policy for the heating process. The cost for handling scrap materials and replacing and reheating new material should be significantly reduced, resulting in a more environmentally friendly, sustainable, and competitive heating process. To optimize the heating process, the outputted billets must always be within the given temperature specification and never exceed the maximum or minimum temperature. The controlling algorithm should adjust the zone voltages individually to even out the existing heating temperature profile. To reduce the speed of material quality degradation due to high temperatures, during hold mode, the temperature of bars inside the furnace should be lowered as much as possible without affecting the restarting time.

\subsection{Digital Twin Assisted Smart Forging using Deep Reinforcement Learning}

\begin{figure*}[t]
    \centering
    \includegraphics[width=0.8\textwidth, ]{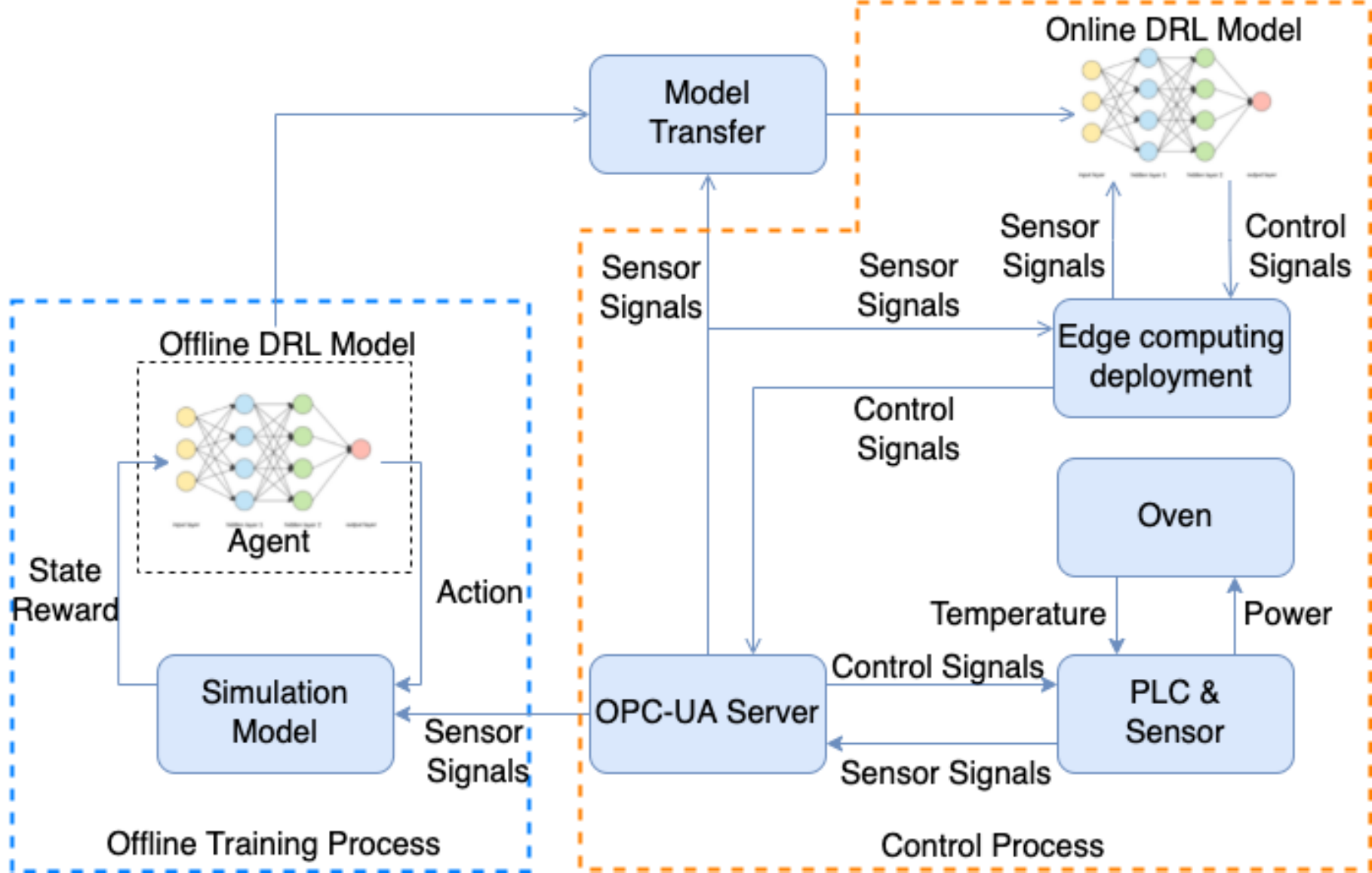}
    \caption{Digital Twin-Based Deep Reinforcement Learning Framework}
    \label{fig:framework}
\end{figure*}

We aim to devise a digital twin-based control framework (DTRL), which is centered around a digital twin of the heating line (see Figure \ref{fig:framework}). The simulation model inside the digital twin of the production process acts as an environment for the agent and simulates the physical dynamics at every time step. In the offline training process (blue dashed box), the offline DRL models (warm holding: $DRL\_wh$, normal production: $DRL\_n$) are trained by the agent, which observes states and rewards by instrumenting the simulator and outputs actions (e.g., adjust power) to the simulation model. The temperatures, positions and speed are mapped into the agent's states. The agent observes the states and the immediate rewards to update the neural networks. The offline models are transferred to the online DRL models ($DRL^\ast\_wh$, $DRL^\ast\_n$) by using e.g., transfer learning. The online DRL models are deployed on an edge compute platform for real-time interaction with the control process to adjust parameters of the heating process for the forging line (e.g., power levels).

In the control process (orange dashed box), the heating oven is controlled by a PLC to steer the heating process by adjusting the powers to the induction coils or changing the speed of conveyor rollers. Sensor signals (observed temperature from the pyrometer sensors, the conveyors' movements, and voltage levels sent to induction coils) are logged by an  OPC-UA server, which can instrument the PLC by overwriting the control policy (currently through a manual operator and a dashboard). The online DRL models are deployed in an edge compute platform, subscribe to the OPC-UA monitoring data, and predict the optimal control signals for the next time step, which are forwarded to the PLC via the OPC-UA server. 

We train the models on the digital twin in this work. Consequently, the digital twin should resemble the conditions of the physical world as good as possible so that the offline learned policy is also a good policy on real data. If there is a mismatch between digital twin and real system, the model needs to be transferred to facilitate the automatic power adjustment to the coils based on real-time data available from the pyrometers. 
This requires drift detection algorithms that use real data and identify, if the model learned by the simulator is not a good match of the actual environment. In that case, the online DRL model can be improved by transfer learning to adjust the optimal policy using observed sensors signals and additional safety controls, which is outside the scope of this paper and left for future work. In the following, we focus on the development of the digital twin and the offline DRL algorithm design in Section \ref{RL_model}.

\subsection{Digital Twin of the Heating Line}
We developed a digital twin of the heating line, which can simulate the process of heating for one or several steel bars by moving them through one or several coils arranged linearly. Each coil is defined by its range and its power. Two positions along the production line define the range of the coil: start position and end position. The simulator assumes that only the segment of the steel bar within the range of a coil is heated with the coil's power. Other segments of the steel bar are not affected by the coil. 

A steel bar is defined by its dimensions, mass, and specific heat coefficient. The digital twin is implemented as a discrete event simulator, where the position of the bar along the production line is updated by the \emph{Movement Manager} for every simulation step. The Movement Manager has two modes of operations:

 \begin{description}
    \item[Normal:] in operation, the steel bars only move forward with a constant configurable speed.

    \item[Warm holding:] during warm holding, the steel bars move forwards and backward with a configurable warm holding speed -- the direction changes at regular intervals.
 \end{description}

\noindent When the steel bar reaches the end of the heating line, the simulator assumes that the bar is moved to the next step in the production. Hence, the simulator will remove the bar. The steel bar history of the movement and its temperature profile is kept for further analysis. The temperature of the steel bar is updated on every time step of the simulator by the \emph{Temperature Manager}. For every segment along the steel bar (within a configurable resolution), the segment is heated if it is within the coil's range. The details of the heating and cooling processes are described in Section \ref{Heating Process Model}. The simulator also provides an interface for algorithms to interact with it using \emph{Controller Manager} class. The Controller Manager continuously provides data to the algorithms and returns the algorithm responses to the simulator. Only the data which can be realistically provided to the algorithms in the real production environment (e.g., the temperature data, steel bar parameters, etc.) are provided. We use that interface to connect the digital twin to our DRL agent to serve as an environment.

\subsection{Heating Process Model}\label{Heating Process Model}
The heating process model in the simulator assumes a one-dimensional bar divided into segments of equal length and uniform temperature. There is no heat exchange between the segments. The total power absorbed $P_{tot}$ over an interval of time $\Delta t = t_{i+1} - t_{i}$ by each segment of length $L$ located under a coil is      
\begin{equation}
    P_{tot} = \frac{m C_V\Delta T}{\Delta t} = kP_{coil} - P_{conv} - P_{rad}.
\end{equation}
On the left-hand side of this equation, $m$ is the mass of the segment, $C_V$ its specific heat, and $\Delta T$ its change in temperature over $\Delta t$. On the right-hand side, $P_{coil}$ is the power of the coil, $k$ the coil$-$bar heat transfer efficiency, and $P_{conv}$ and $P_{rad}$ the power loss (i.e., cooling) due to convection and radiation, respectively.

In this work, the heat transfer efficiency $k$, which changes drastically at the Curie temperature $T_{C}$ ($\approx 800 ^\circ C$  for steel), is determined empirically from the process data. Below $T_{C}$, $k \approx 90 \%$, whereas it drops to as low as $20 \%$ above $T_{C}$.    
In our model, the convective cooling term is given by
\begin{equation}
    P_{conv} =  \pi D L \cdot 1.86 (T_{s}-T_{a})^{1.3}
\end{equation}
where $D$ is the diameter of the segment, $T_{s}$ is surface temperature, and $T_{a}$ is the ambient temperature. The factor $1.86$ and exponent $1.33$ are engineering estimates taken from literature\cite{Rudnev2017induction}. Only free convective cooling is assumed; a negligible amount of forced convective cooling occurs since the bar moves with slow speed ($\approx4$ cm/s) and since there are no other mechanisms, e.g., fans, to force heat away from the bar before shearing.  A rough estimate of radiative cooling is
\begin{equation}
    P_{rad} = \sigma \epsilon ((T_{s} + 273)^{4} - (T_{a} + 273)^{4})
\end{equation}
where $\sigma$ is the Stefan Boltzmann constant, and $\epsilon$ is the emissivity of the bar. $T_{s}$ and $T_{a}$ are expressed in $^\circ C$.

\subsection{Deep Reinforcement Learning Model and Agent}\label{RL_model}
In our approach, the agent needs to find the optimal control policy using the digital twin as an environment during the offline training phase.
We apply the PPO algorithm\cite{schulman2017proximal}, which is an on-policy actor-critic algorithm that incorporates a replay buffer and neural networks. The replay buffer stores trajectories experienced by the PPO agent when interacting with the environment. The policy network predicts the stochastic policy that is usually parameterized as a gaussian distribution with mean and variance, and the critic network predicts the states' values. We calculate the returns and advantage estimates with the data stored in the replay buffer at the end of a trajectory. A central feature of the PPO algorithm is the objective function of the policy network. It calculates the most significant improvement step that can be made without causing performance collapse. The main objective proposed by \cite{schulman2017proximal} is the following:
\begin{equation}
    L^{CLIP}(\theta) = \hat{\mathbb{E}_{t}}\left[ min\left(r_t\left(\theta\right)\hat{A_t},clip\left(r_t\left(\theta\right), 1-\epsilon, 1 + \epsilon\right)\hat{A_t}\right)\right]
\end{equation}
where $r_{t}\left(\theta\right)=\frac{\pi_\theta\left(a_t|s_t\right)}{\pi_{\theta_{old}}\left(a_t|s_t\right)}$, the probability ratio, $\pi_\theta\left(a_t|s_t\right)$ and $\pi_{\theta_{old}}\left(a_t|s_t\right)$ denote the probabilities of choosing action $a_t$ given a state $s_t$ under the new policy $\pi_\theta$ and the old policy $\pi_{\theta_{old}}$, separately. The $\hat{A_t}$ is the advantage estimate. The first term inside the min is the objective without a constraint. The second term, $clip\left(r_t\left(\theta\right), 1-\epsilon, 1 + \epsilon\right)\hat{A_t}$, modifies the objective by clipping the probability ratio within the interval $\left[1-\epsilon, 1+\epsilon\right]$, where $\epsilon$ is a hyperparameter. Finally, we take the minimum of the clipped and unclipped objective, so the final objective is  a lower bound on the unclipped objective.

The PPO agent interacts with the digital twin using a markov decision process (MDP)-based approach. It receives a state vector $s_{t}$ and outputs an action vector $a_{t}$ at time step $t$. The simulation environment is updated to a new state $s_{t+1}$ with the action vector $a_{t}$, and gives a reward $r_{t+1}$. The PPO agent observes the new state $s_{t+1}$ and predicts a new action $a_{t+1}$. Next, we describe the states, actions and reward functions that our PPO agent uses: 

\emph{States:} One steel bar is divided into $n$ temperature elements, where temperature element $i$ at time step $t$ is denoted as $T_{t,i}$. The state vector includes: temperature elements of the steel bar, bar position, time step. Let's denote the position at the head of steel bar as $Pos_{head}$, the position at the tail of steel bar as $P_{tail}$, time step as $t$. The state vector is: $S_t = [T_{t,1}, T_{t,2}...T_{t,n}, Pos_{head}, Pos_{tail}, t]^T$.

\emph{Actions:} At time step t, the actions are suggested power from Zone 3 to Zone 5. Then forming the action vector at time step t: $ a_t = [P_{t,3},P_{t,4},P_{t,5}]$,
where the power for each zone is limited by  $P_{max}$: 
$0 < P_{t,3},P_{t,4},P_{t,5} < P_{max}$.
We do not consider the powers at Zone 1 and Zone 2 as learnable actions for two reasons: 1. There are only two power levels at Zone 1, and it can not be treated as continuous actions as other zones. 2. The powers at Zone 1 and Zone 2 are much higher than in other zones. In practice, we find our agent performs better if the powers at Zone 1 and Zone 2 are fixed, which reduces the search space.

\emph{Reward:} Our reward function consists of three parts: $R_{t}=R_{t,even}+R_{t, heat}+R_{t, move}$. $R_{t,even}$ penalizes the unevenness of the temperature distribution over the steel bar, $R_{t, heat}$ penalizes the overheat using soft constraints and $R_{t, move}$  encourages the bar to move towards the shearing machine.
The $R_{t,even}$ is the negative logarithm of mean absolute errors of all the temperature elements over the steel with base 10, i.e., $R_{t, even} = - \log_{10}(1/n\cdot \sum_{i=0}^n |T_{t,i} - T_{target}|)$, where $T_{target}$ is the target temperature at each zone. $R_{t, heat}$ is a negative constant value, and $R_{t, move}$ is the scaled position of the bar's head. 

In this paper, we are targeting to optimize the process for a bar of size 4m. Our evaluations showed that the soft constraints in the reward function perform well to prevent the bar from overheating. However, in practice, we need to give hard constraints for bars longer than 10m. When any part of the bar's temperature is close to the maximum temperature, we can, for example, mask the output power from the DRL agent by turning off this zone's power. This strategy is effective to assure the DRL model outputs safe actions for the longer bar. 

We extend the digital twin for connecting it to the OpenAI Gym environment \cite{Brockman2016gym}, which we used for implementing the agent. We normalize the simulator's temperature, position, and time steps when feeding it to the PPO agent as the state vector. We also map the action vector predicted by the policy network to the actual powers, and update the simulator's temperatures and positions with the actual powers at the next time step. For training the warm holding model, we optimize the movement using constant power and find the best movement pattern that results in the lowest final temperature unevenness using grid search. Fixing the movement pattern reduces the searching space and helps the PPO algorithm to converge.

\section{Evaluation}\label{Evaluation}
Our experimental evaluation aims to understand whether the heating process can be optimized by using the DRL algorithm in the simulation environment and whether the solution meets the temperature requirements. We create two separate DRL models: model $DRL\_n$ aims at optimizing the normal production process and model $DRL\_{wh}$ aims at optimizing the warm holding process and the resumed normal production process after the warm holding is finished. The two models are evaluated separately. For deployment in the real system, the DTRL framework can use $DRL\_n$ to control the power and optimize the final temperature during normal production mode. When the operator switches on the warm holding, the framework can switch to model $DRL\_{wh}$. Note that an evaluation of the algorithm with real production data together with its deployment in an edge device is subject to future work.

\subsection{Evaluation Setup}
The DTRL framework uses the following parameters to train the DRL models: 
\begin{itemize}

\item \emph{Simulation model parameters}: we use a bar length of 4m, a fixed roller speed and a band length of $30 m$. The powers at zone 1 and zone 2 are set to $2MW$ and $4.2 MW$. The output power is limited to $0.4MW$ at zone 3, zone 4, and zone 5 in both modes. The maximum temperature of the bar should not exceed 1100 $^{\circ}C$. During the warm holding mode, the grid search provided the following empirical parameters: warm holding time $540s$, the bar moves backward or forwards for $64s$ in the first, fourth, fifth, eighth turns and moves for $60s$ in the second, third, sixth, seventh turns.  

\item \emph{PPO hyperparameters}: We use default settings from the OpenAI Spinning Up\footnotemark, clip\_ratio of 0.02 and learning rate of 0.0001.
\footnotetext{https://spinningup.openai.com/en/latest/algorithms/ppo.html}
\item \emph{Weights of the reward function}: We use $1$ for the weights of $R_{t,even}, R_{t,move}$. We clip the $R_{t,even}$ from $-2$ to $1$, the range of $R_{t,move}$ is from 0 to 3, the range of $R_{t,heat}$ is constant value of $-5$. With this setting, the total reward range is from $-7$ to $4$. We did not scale the reward range to [-1, 1] since the reward is small enough.
\end{itemize}

\subsection{Results}\label{results}
\begin{figure}
    \centering
    \begin{minipage}[b]{0.4\textwidth}
    \includegraphics[width=\textwidth]{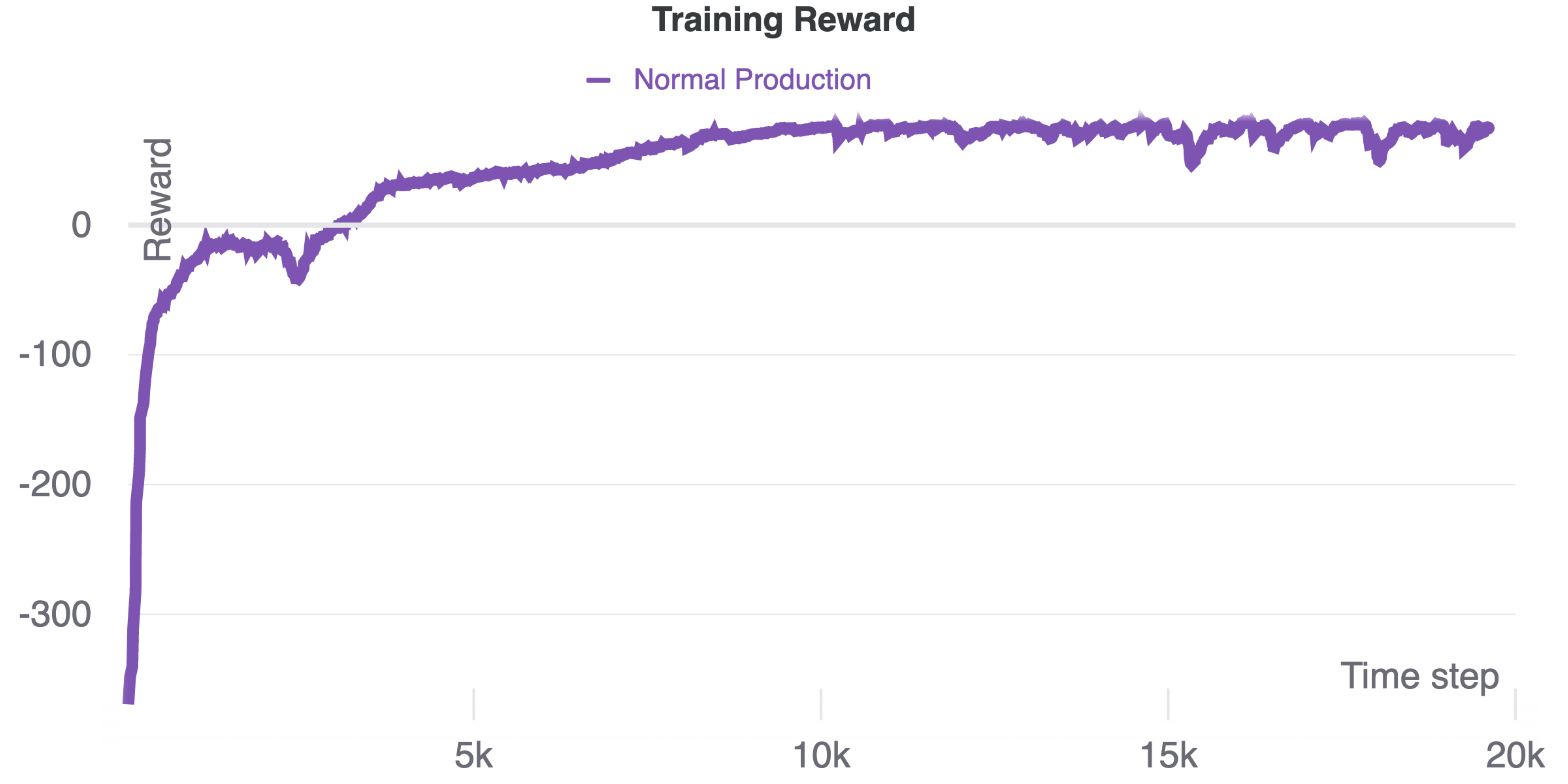}
    \caption{Training curves in normal production mode}
    \label{fig:reward_normal}
    \end{minipage}
    \hfill
    \centering
    \begin{minipage}[b]{0.4\textwidth}
    \includegraphics[width=\textwidth]{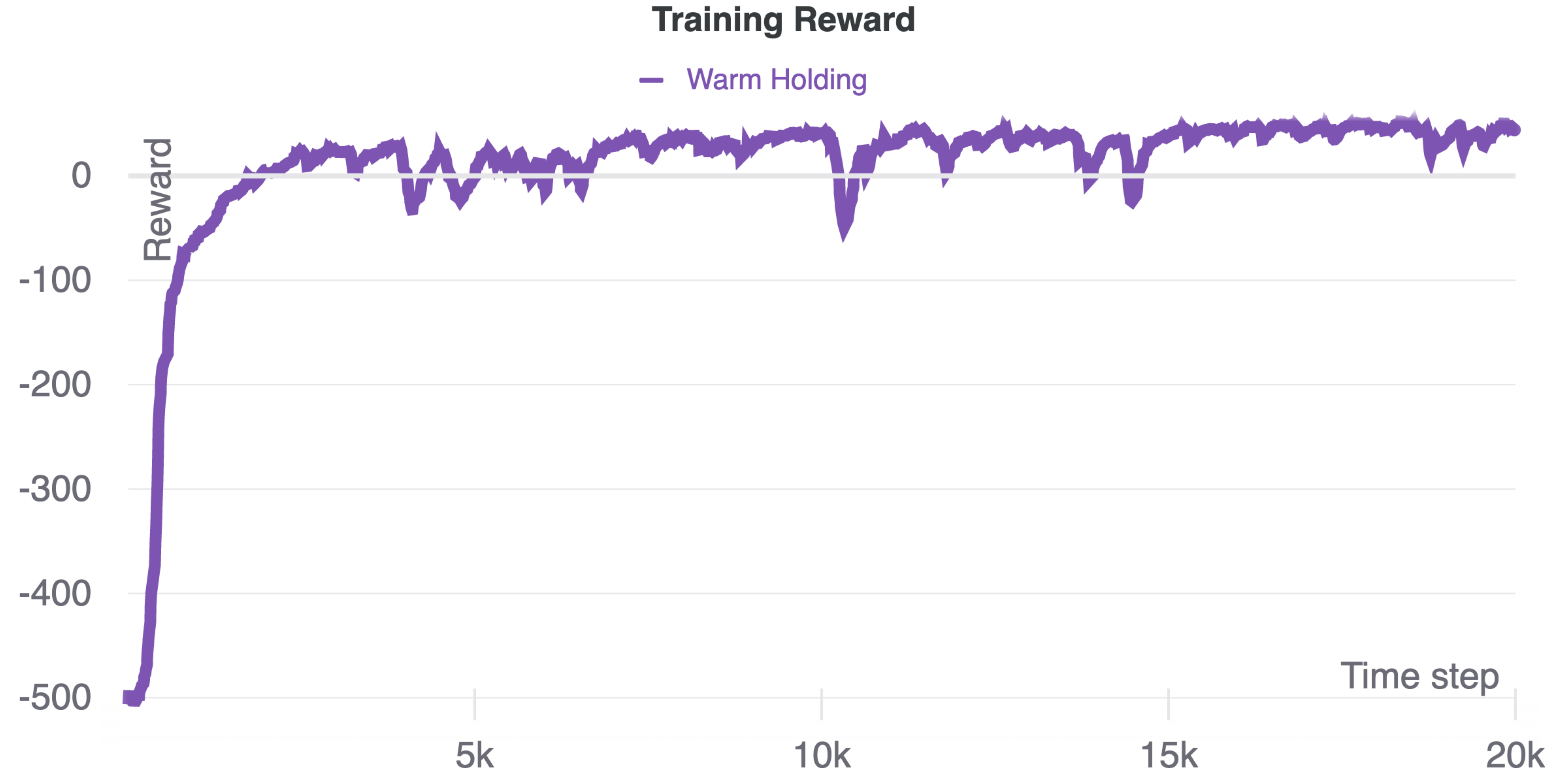}
    \caption{Training curves in warm holding mode}
    \label{fig:reward_wh}
    \end{minipage}
\end{figure}

We first train the offline models separately using the digital twin. 
Figure \ref{fig:reward_normal} and Figure \ref{fig:reward_wh} show the average of recent 100 episodic rewards during the training phase for the normal production model $DRL\_n$  and the warm holding model $DRL\_{wh}$, respectively. We can see a rapid increase in the reward at the beginning of the training process because the initial temperature distribution is already close to the required temperature range, which benefits from a narrow power range setting and movement optimization process (for the warm holding model). The low changes of rewards, in the end, demonstrate the convergence of the training. The mean episode rewards increase from -380 to 80 for $DRL\_n$ and from -500 to 30 for $DRL\_wh$. This is determined by the reward weights and the simulation model parameters. The reward curve shows more instability for the warm holding model because of the higher complexity of the process, where the steel bar's temperature has more chance to get overheated.

\begin{figure}
    \centering
    \begin{minipage}[b]{0.4\textwidth}
    \includegraphics[width=\textwidth]{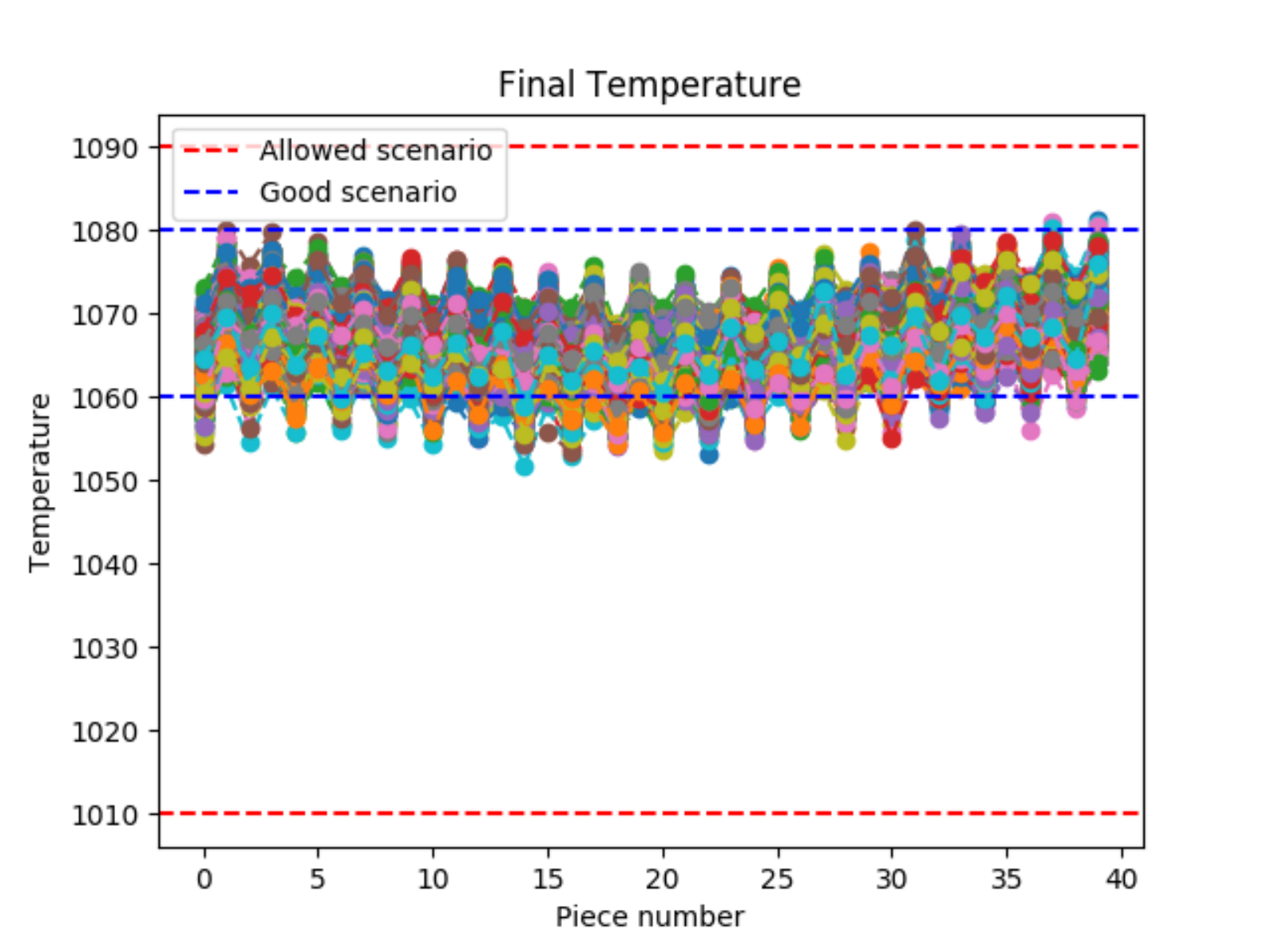}
    \caption{Final temperature in normal production mode}
    \label{fig:final_normal}
    \end{minipage}
    \hfill
    \centering
    \begin{minipage}[b]{0.4\textwidth}
    \includegraphics[width=\textwidth]{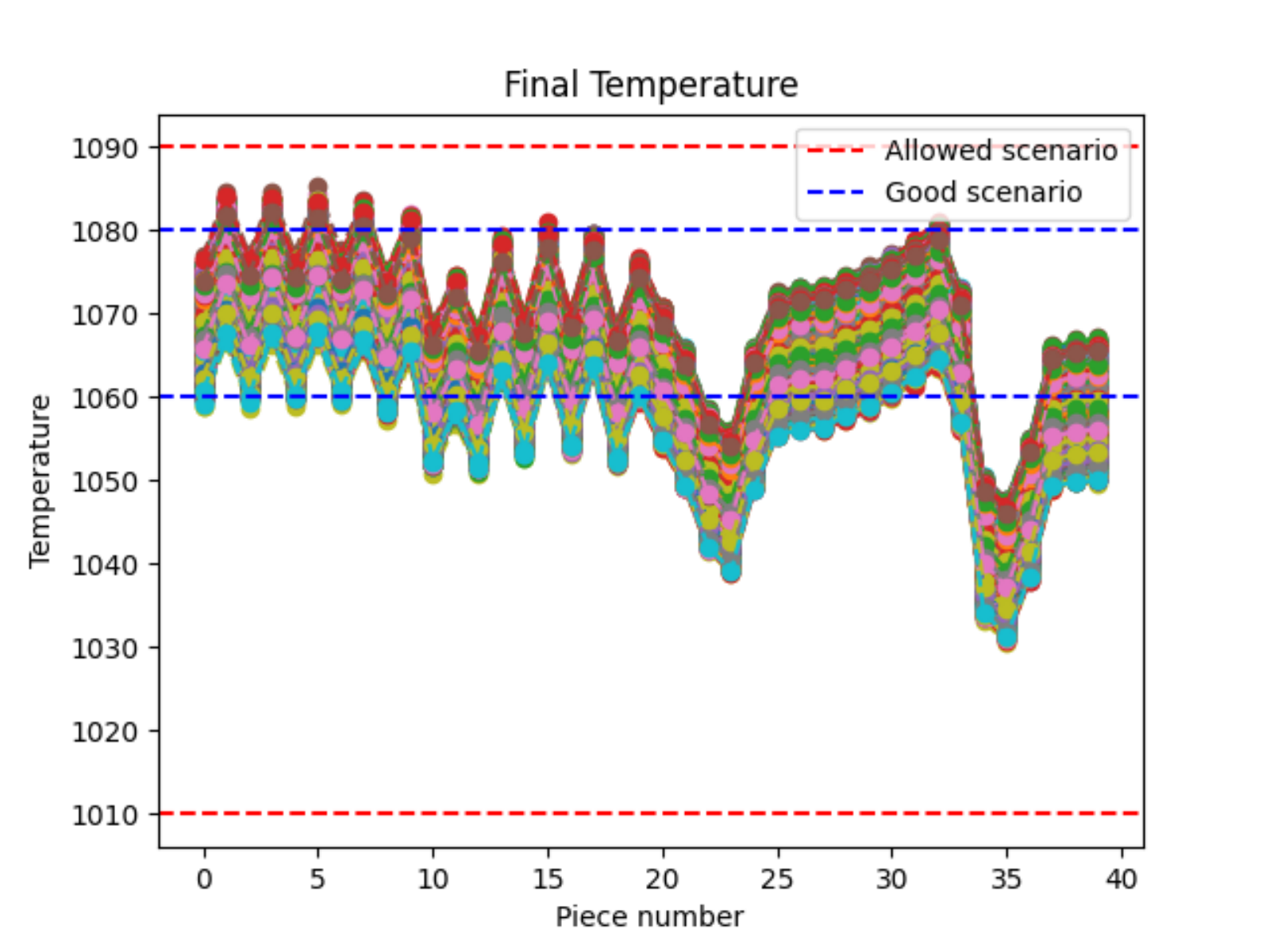}
    \caption{Final temperature in warm holding mode}
    \label{fig:final_wh}
    \end{minipage}
\end{figure}

After training the offline models of $DRL\_wh$ and $DRL\_n$, we evaluate them with a stochastic policy by running one thousand episodes each to test if an anomaly occurs. Figure \ref{fig:final_normal} and Figure \ref{fig:final_wh} show the final temperature for each piece when leaving the heating process both for normal process and warm holding mode.
The x-axis represents the piece index $i$ of the steel bar, and the y-axis plots the temperature distribution of the final piece after leaving the heating process over the 1000-time episodes. The area between the horizontal dash line in red color is the temperature range required in the production ($1010 ^\circ C - 1090^\circ C$), and the area between the horizontal dash line in blue color is the desired temperature range ($1060 ^\circ C - 1080^\circ C$) for the optimal scenario.  Figure \ref{fig:final_normal} shows that most of the pieces' temperatures during normal production mode are within the desired range, and the temperature over the steel bar is smoothly distributed. For the warm holding model, Figure \ref{fig:final_wh} shows that about 75\% of the pieces' temperatures are in the target range, and about 25\% of the pieces' temperatures are colder than desired but still meet the production requirements. However, during all runs, we never observed any overheating during the whole process duration and the temperatures always stayed below $1100 ^\circ C$.

\subsection{Discussion}
Comparing the two DRL models, the final temperature predicted by $DRL\_n$ is more evenly distributed than $DRL\_wh$. The reason is that $DRL\_wh$ is more dependent on extrinsic factors, e.g., the initial temperature distribution over the steel bar and the initial position when warm holding is initiated, movement pattern, and the given power range. In real production, the warm holding can start at any time, which makes it challenging to get an accurate initial temperature distribution when warm holding is activated. Moreover, our experiment found that the movement pattern during warm holding and power settings to the induction coils also drastically affected the DRL model's performance. In practice, these settings should correspond to the real production, or we should guarantee that the problem has a solution with those settings. Finally, the limited action dimensions also hinder the $DRL\_wh$ model from getting an optimal solution. This impact can be reduced by different predefined movement patterns that are optimized before the DRL model training.

\section{Conclusion and future work}\label{Conclusion}
In this paper, we propose a digital twin based deep reinforcement learning approach to optimize the heating process of a forging line, which is currently partly automated using pre-made recipes and highly skilled manual operations. In our approach, we aim to reduce the scrapped materials produced in the warm holding process by using AI-based methods aiming towards zero defect manufacturing. To achieve this, we developed a digital twin that modeled the heating process and adapted it to use it as an environment for training a deep reinforcement learning agent. We use the PPO algorithm and we train models for normal production and warm holding modes. Our evaluation shows that the final temperature achieved by our approach meets the production's requirement. 

Future work will focus on improving the robustness and generalization of the trained models to different warm holding patterns. Another line of research will study how the DTRL framework can be deployed automatically in a real-time production platform, including transfer learning to adjust the model to match the real production line environment. That needs to include methods for online drift detection and automatic retraining of the model during production, which requires correlating the simulator heating model with the real conditions.

\section*{Acknowledgments}

Parts of this work has been funded by Vinnova, Sweden's Innovation Agency, through the FFI Project SmartForge\footnote{Rek:2020-02835} - Sustainable production through AI controlled forging oven.

\bibliographystyle{vancouver}
\bibliography{references}
\end{document}